\documentclass[journal]{IEEEtran}

\usepackage{multirow}
\usepackage{graphicx}
\usepackage{cite}
\usepackage{subcaption}
\usepackage{xcolor}
\usepackage{placeins}
\usepackage{amsmath}
\usepackage{colortbl}
\usepackage[subpreambles]{standalone}

\usepackage{threeparttable}%
\usepackage{array}
\newcolumntype{L}[1]{>{\raggedright\let\newline\\\arraybackslash\hspace{0pt}}m{#1}}
\newcolumntype{C}[1]{>{\centering\let\newline\\\arraybackslash\hspace{0pt}}m{#1}}
\newcolumntype{R}[1]{>{\raggedleft\let\newline\\\arraybackslash\hspace{0pt}}m{#1}}

\begin{document}
\title{ConvNet-Based Localization of Anatomical Structures in 3D Medical Images}
\author{
	Bob~D.~de~Vos, 
	Jelmer~M.~Wolterink,
	Pim A. de Jong,
	Tim Leiner,
	Max~A.~Viergever,
	Ivana~I\v{s}gum\\

	\thanks{This paper was submitted in December 2016 and accepted in February 2017 as: 
	Bob D. de Vos, Jelmer M. Wolterink, Max A. Viergever and Ivana I\v{s}gum. 
	ConvNet-Based Localization of Anatomical Structures in 3D Medical Images. IEEE Transactions on Medical Imaging. PP. DOI: 10.1109/TMI.2017.2673121}
	\thanks{
		This study was funded by the Netherlands Organization for Scientific Research (NWO)/ Foundation for Technology Sciences (STW); Project 12726. We gratefully acknowledge the support of NVIDIA Corporation with the donation of the Tesla K40 GPU used for this research. The authors thank the National Cancer Institute for access to NCI's data collected by the National Lung Screening Trial. The statements contained herein are solely those of the authors and do not represent or imply concurrence or endorsement by NCI.}

	\thanks{Bob D. de Vos, Jelmer M. Wolterink, Max A. Viergever and Ivana I\v{s}gum are with the Image Sciences Institute, University Medical Center Utrecht, Utrecht, The Netherlands. Pim A. de Jong and Tim Leiner are with the Department of Radiology, University Medical Center Utrecht, the Netherlands}
}

\maketitle

\begin{abstract}
Localization of anatomical structures is a prerequisite for many tasks in medical image analysis.  We propose a method for automatic localization of one or more anatomical structures in 3D medical images through detection of their presence in 2D image slices using a convolutional neural network (ConvNet).

A single ConvNet is trained to detect presence of the anatomical structure of interest in axial, coronal, and sagittal slices extracted from a 3D image. To allow the ConvNet to analyze slices of different sizes, spatial pyramid pooling is applied. After detection, 3D bounding boxes are created by combining the output of the ConvNet in all slices. 

In the experiments 200 chest CT, 100 cardiac CT angiography (CTA), and 100 abdomen CT scans were used. The heart, ascending aorta, aortic arch, and descending aorta were localized in chest CT scans, the left cardiac ventricle in cardiac CTA scans, and the liver in abdomen CT scans. Localization was evaluated using the distances between automatically and manually defined reference bounding box centroids and walls. 

The best results were achieved in localization of structures with clearly defined boundaries (e.g. aortic arch) and the worst when the structure boundary was not clearly visible (e.g. liver). The method was more robust and accurate in localization multiple structures.
\end{abstract}


\IEEEpeerreviewmaketitle
\section{Introduction}
\IEEEPARstart{L}{ocalization} of anatomical structures is a prerequisite for many tasks in medical image analysis. Localization may be a simple task for human experts, but it can be a challenge for automatic methods, which are susceptible to variation in medical images caused by differences in image acquisition, anatomy, and pathology among subjects. Nonetheless, automatic localization methods have been used as a starting point for automatic image segmentation and analysis (e.g.~\cite{seifert2009,gauriau2015,okada2015,zreik2016-3002,Wolterink2016123,lessmann2016-2955}), for categorizing images~\cite{roth_anatomy-specific_2015}, and for retrieval of (parts of) medical images from PACS databases (e.g.~\cite{gueld2002,criminisi2013}).

In literature, \textit{detection} and \textit{localization} are often used interchangeably. In this paper, we use two strict definitions: we define \textit{detection} as determining presence of a target object in an image and we define \textit{localization} as determining a region in an image that contains the target object. This differs from \textit{segmentation} methods that delineate a target object as accurately as possible.

Very few dedicated methods for automatic localization of organs or anatomical structures have been proposed (e.g.\cite{zheng2009,zhou2012,criminisi2013,ghesu2016,devos2016,lu2016}).
These methods can be separated in those using atlas-based registration and those using machine learning. Atlas based registration is generally used for segmentation, but it can also be applied for localization. In registration, atlases, i.e. images with known locations of anatomical target structures, are registered to an image with unknown locations of anatomical target structures. By transforming these images to a common space, anatomical structures can be roughly outlined\cite{rohlfing2004, vanderlijn2008, isgum2009, kirisli2010, ranjan2011}. In this fashion one or multiple anatomical structures can be localized in parallel. Atlas-based registration has proven to be a robust technique, but it typically requires considerable parameter tuning, and is considered to be slow. Ideally, a localization method should be robust, fast, and easy to apply. Therefore, dedicated approaches have been developed that try to meet these requirements. 

Zheng et al.~\cite{zheng2007} introduced a dedicated localization method employing marginal space learning (MSL). This learning algorithm is used in \cite{zheng2007} to predict coordinates describing a bounding box around single anatomical target structures in either 2D or 3D medical images. A bounding box is determined in three consecutive stages. The first stage determines the location of the bounding box, the second stage determines its orientation, and the third stage determines its size. In this way, every stage reduces the search space for the subsequent parameter search while still allowing adjustment of the parameters found in previous stages. In every stage candidate bounding boxes are generated and the optimal box is chosen with a probabilistic boosting tree. The method was evaluated with localization of the left ventricle in 2D MRI images, and with localization of the left ventricle of the heart and liver in 3D CT volumes~\cite{zheng2009}. 

Ghesu et al.\cite{ghesu2016} replaced the probabilistic boosting tree in MSL with sparse adaptive deep neural networks and cascaded filtering. These networks are fully-connected, highly sparse, and are very efficient in analysis of 3D images. The approach was coined marginal space deep learning (MSDL). Like in MSL, a bounding box is determined around a target structure in three consecutive stages. Unlike a boosting tree, the neural networks in MSDL allow automatic design of features. The method has been evaluated with aortic valve localization in 3D ultrasound volumes and improves upon MSL. However, like in MSL, MSDL can only localize one anatomical target structure at a time. Localizing multiple target structures would increase localization time linearly.

A method that allows localization of multiple anatomical target structures was proposed by Criminisi et al.\cite{criminisi2013}. In this work the localization problem is posed as a multivariate regression problem, and solved by a regression forest. The regression forest determines the position of the six bounding box walls for the target  structures based on texture features calculated from input voxels~\cite{shotton_textonboost_2007}. The method was evaluated with localization of 26 organs in full-body CT. Gauriau et al.~\cite{gauriau2015} improved upon this work by cascading the initial regression forest with dedicated per-target forests. The addition to the method reduced localization errors, but it increased localization time.

Zhou et al.~\cite{zhou2012} localized organs in 3D scans by initially creating 2D bounding boxes in image slices from three orthogonal planes. 2D bounding boxes were determined with AdaBoost and an ensemble of stump classifiers exploiting Haar-like features. Thereafter, these 2D bounding boxes were combined into 3D bounding boxes by majority voting. The method was evaluated with localization of five different organs in CT images. The method is slower than the previously mentioned methods and only localized one organ at a time.

 We propose a method that meets the requirements of robustness, ease of use, and short processing time. The proposed method localizes a single anatomical structure or multiple structures in 3D medical images by mimicking a human observer: the method poses \textit{3D localization} as a \textit{2D detection problem}. Anatomical structures are localized in 3D images by first determining their presence in 2D image slices. A convolutional neural network (ConvNet) detects presence or absence of the anatomical target structures in each of the orthogonal viewing planes. Thereafter, a bounding box in 3D is obtained by combining the ConvNet outputs for all axial, coronal and sagittal slices. 
 This idea was previously proposed in our preliminary work \cite{devos2016} and a very recent work by Lu et al. \cite{lu2016}. In \cite{devos2016} single anatomical target structures (the heart, aortic arch, and descending aorta) were localized by three independent ConvNets, where each ConvNet evaluated image slices of a single image plane (axial, coronal, or sagittal). The three ConvNets were trained separately and did not share weights. In \cite{lu2016} pixels from the 2D image slices, as well as the 2D image slices themselves, were classified as being present or absent in a bounding box containing the organ of interest: the right kidney. Pixels were classified by a ConvNet with a small receptive field to capture the local context, and 2D image slices were classified by a ConvNet with a large receptive field to capture the global context. Subsequently, the outputs of the ConvNets for the image slices were combined by summation. The largest connected component determined the bounding box around the organ of interest.
 
The here proposed method employs a single ConvNet to detect presence of anatomical target structures in 2D image slices extracted from all three image planes of a 3D medical image. Image slices are \textit{independently} evaluated by the ConvNet, which outputs probabilities for presence of one or multiple target structures. The number of target structures determines the number of output nodes of the ConvNet. For each target structure, one output node predicts its presence and a linked output node predicts its absence. The linked output nodes enable the use of a softmax output~\cite{bridle1990}, while still allowing multilabel classification. Furthermore, in \cite{devos2016} the ConvNets required input images of fixed sizes. Given that input image slices are typically of variable size, the input image slices were cropped or padded. To allow localization in images of different sizes, thereby avoiding the need for cropping or padding, spatial pyramid pooling~\cite{he_spatial_2015} is used in this work. Finally, the evaluation has been substantially extended. The dataset has been increased from one to three different sets where localization of six different anatomical structures has been performed. Moreover, performance of the automatic localization has been compared with the performance of a second observer.

 \begin{figure}[t]
	\begin{tabular}{C{.29\columnwidth}C{.29\columnwidth}C{.29\columnwidth}}
		\footnotesize\textsf{Axial} & \footnotesize\textsf{Coronal} & \footnotesize\textsf{Sagittal}
	\end{tabular}
	\begin{subfigure}[c]{\columnwidth}
		{\includegraphics[width=\textwidth, trim=40 150 40 140, clip]{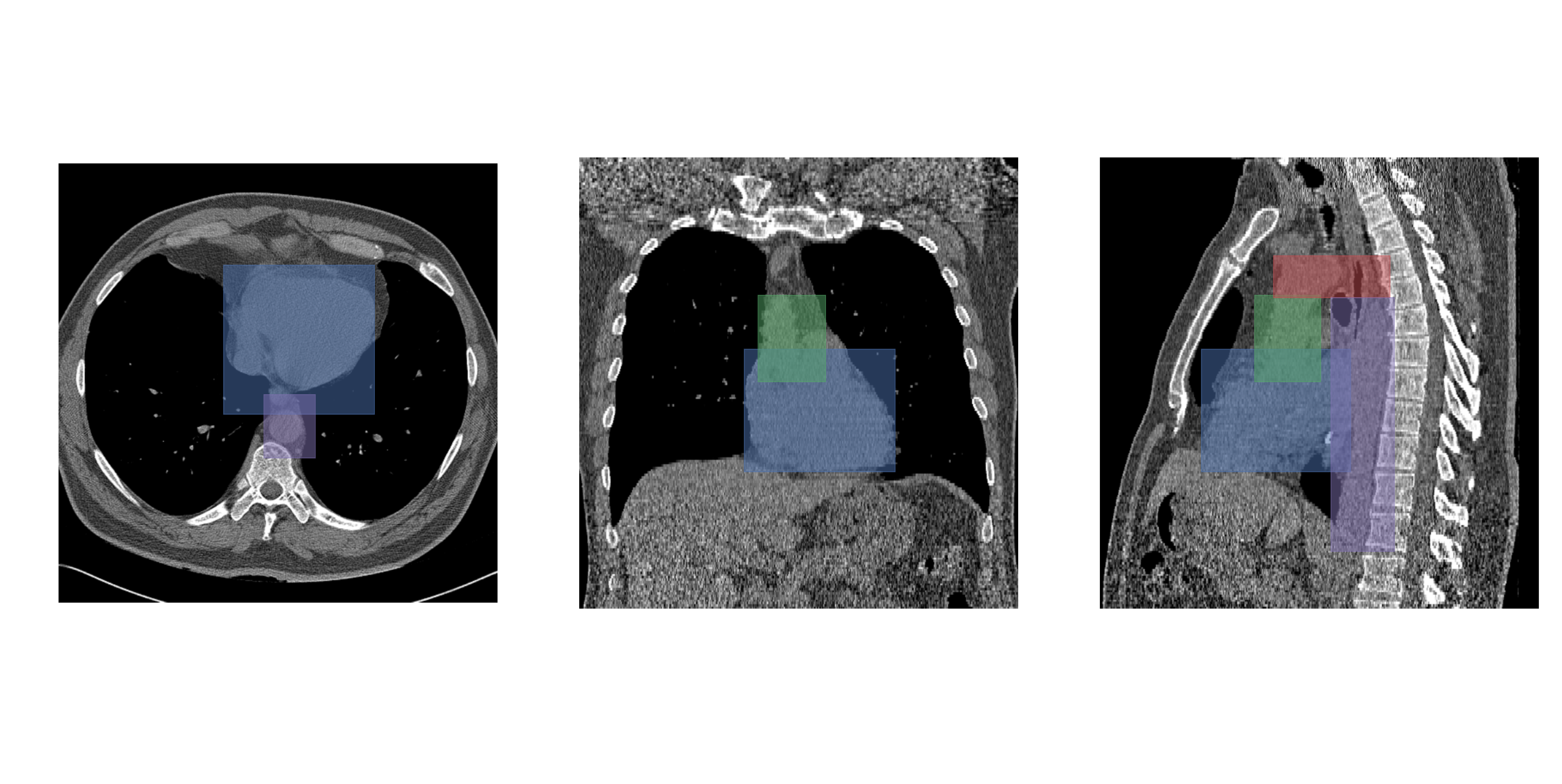}}
	\end{subfigure}%
	\\
	\begin{subfigure}[c]{\columnwidth}
		{\includegraphics[width=\textwidth, trim=40 150 40 140, clip]{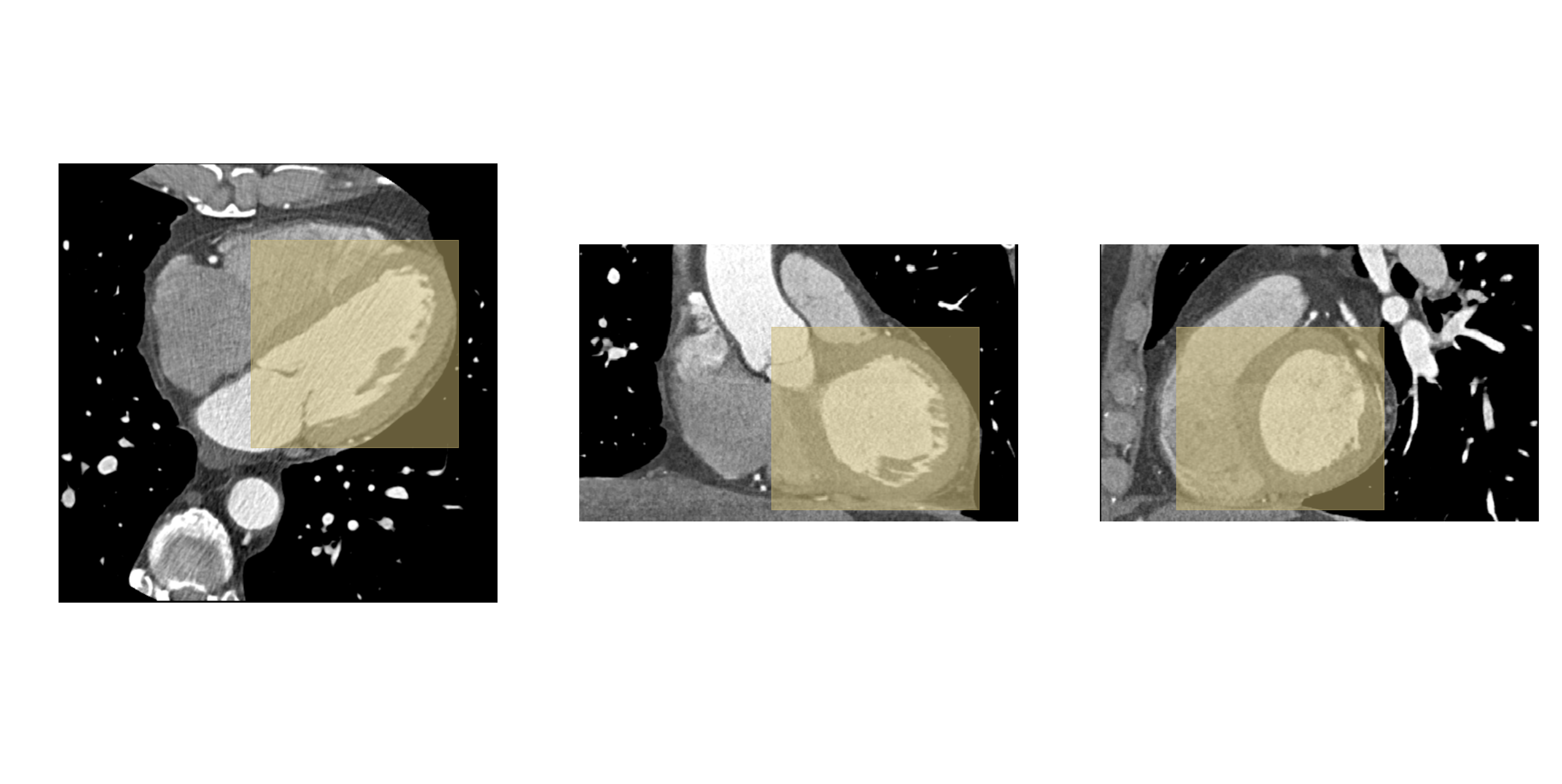}}
	\end{subfigure}%
	\\
	\begin{subfigure}[c]{\columnwidth}
		{\includegraphics[width=\textwidth, trim=20 100 20 70, clip]{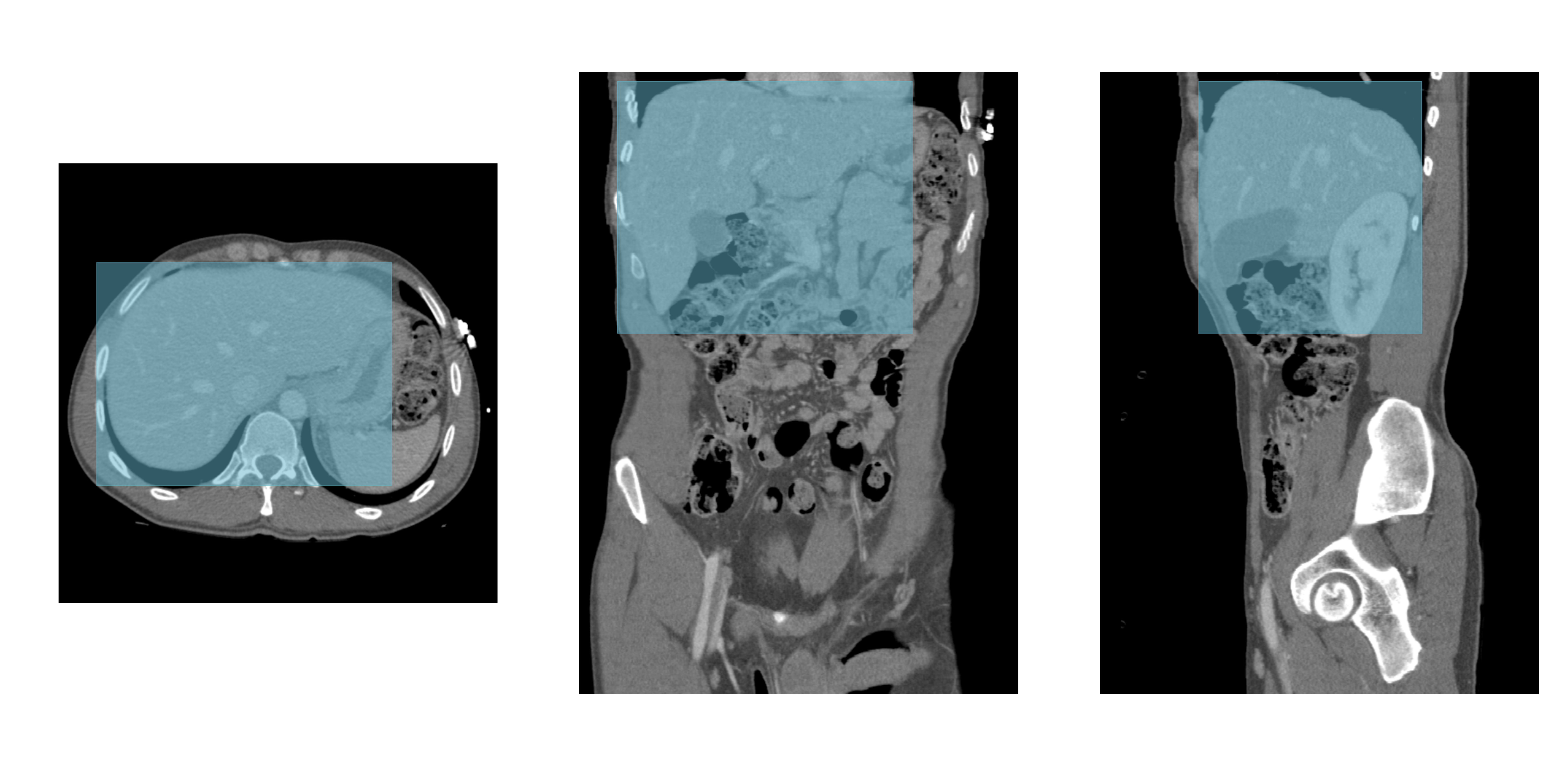}}
	\end{subfigure}%
	\\
	\caption{Three scans from the datasets used in this work are shown in the rows: chest CT (top), cardiac CTA (middle), and abdomen CT (bottom). For every scan image slices extracted from the three orthogonal viewing planes are shown: axial (left), coronal (middle), and sagittal (right). Each image shows one or more rectangular masks indicating reference 3D bounding boxes around all annotated anatomical structures visible in the image: in chest CT the heart (blue), ascending aorta (green), aortic arch (red), and descending aorta (purple); in cardiac CTA the left cardiac ventricle (yellow); and in abdomen CT the liver (light blue).}
	\label{fig:data}
\end{figure}
\section{Materials}
\label{sec:materials}
\begin{table*}[ht]
	\centering
	\caption{Overview of population of the chest CT, cardiac CTA, and abdomen CT scans used in the study. In addition, scanner, acquisition and reconstruction parameters are listed. }
	\label{tab:data}
	\begin{tabular}{llccc}
		\multicolumn{2}{l}{Subject}&Chest CT&Cardiac CTA&Abdomen CT\\
		\hline
		&Age (yrs) & 55--74&19--84&26--85\\
		&Nr. (men/women) & 200 (133/67)& 100 (81/19)& 100 (54/46)\\
		\\
		\multicolumn{5}{l}{Scanner}\\
		\hline
		&Manufacturers                              & GE, Philips, Siemens, Toshiba & Philips & Philips\\
		&Nr. models                  & 10                                & 1 & 5\\
		&Nr. scanners & 35&1&6\\
		\\
		\multicolumn{5}{l}{Acquisition}\\
		\hline
		&Tube voltage (kVp)     & 120, 140  &\phantom{0}80, 100, 120&100, 120, 140\\
		&Tube current  (mAs)         & \phantom{0}30--160 &210--600&\phantom{0}40--410\\
		\\
		\multicolumn{5}{l}{Reconstruction}\\
		\hline
		&In plane voxel size (mm) & 0.50--0.86&0.29--0.49&0.55--0.98\\
		&Slice thickness (mm)           & 1.00--3.20                       &0.90&0.90--2.00\\
		&Slice increment (mm)            & 1.00--2.50                  &0.45&0.45--1.00\\
	\end{tabular}
\end{table*}

\subsection{Data}
In this work, three different sets of CT images were used (see Figure~\ref{fig:data}): chest CT scans,  cardiac CT angiography (CTA) and  CT scans of the abdomen.  A detailed description of acquisition parameters of each dataset is listed in Table~\ref{tab:data}. The method was evaluated with localization of six anatomical structures in these sets. The heart, ascending aorta, aortic arch, and descending aorta were localized in chest CT scans; the left ventricle of the heart in cardiac CTA; and the liver in abdomen CT.

\subsubsection{Low-Dose Chest CT}
A set of 200  baseline scans were randomly selected from a set of 6,000 available baseline scans of the National Lung Screening Trial (NLST)\cite{nlst2011}. All scans were acquired during inspiratory breath-hold in supine position with the arms elevated above the head and included the outer rib margin at the widest patient dimension. The selected scans were acquired on 35 different scanners, without contrast enhancement, with various resolution, and were reconstructed with various reconstruction kernels. 

\subsubsection{Cardiac CT Angiography}
A set of 100 consecutively scanned cardiac CTA scans from the University Medical Center Utrecht (Utrecht, the Netherlands) were retrospectively collected. Informed consent was waived by the local medical ethics committee. Scans were made from the carina to the cardiac apex, excluding the lungs from the field of view as much as possible. All scans were made on one 256-detector row scanner (Philips Brilliance iCT, Philips Medical, Best, The Netherlands) with ECG-triggering and contrast enhancement. 

\subsubsection{Abdomen CT}
A set of 100 consecutively scanned abdomen trauma CT scans from the University Medical Center Utrecht (Utrecht, the Netherlands) were retrospectively collected. Informed consent was waived by the local medical ethics committee. Scans were acquired with six different CT scanners with 16, 64, or 256 detector rows (Philips Medical, Best, The Netherlands). Scans were made with the whole abdomen of the patient in the field of view. The scans were acquired with various tube voltage, various tube current, and various phases of contrast enhancement. Images were reconstructed with two different reconstruction kernels, and with various reconstruction parameters.

\subsection{Reference Standard}
\begin{figure}[h]
	\centering
	\includegraphics[trim=0 0 0 0, clip, width=.8\columnwidth]{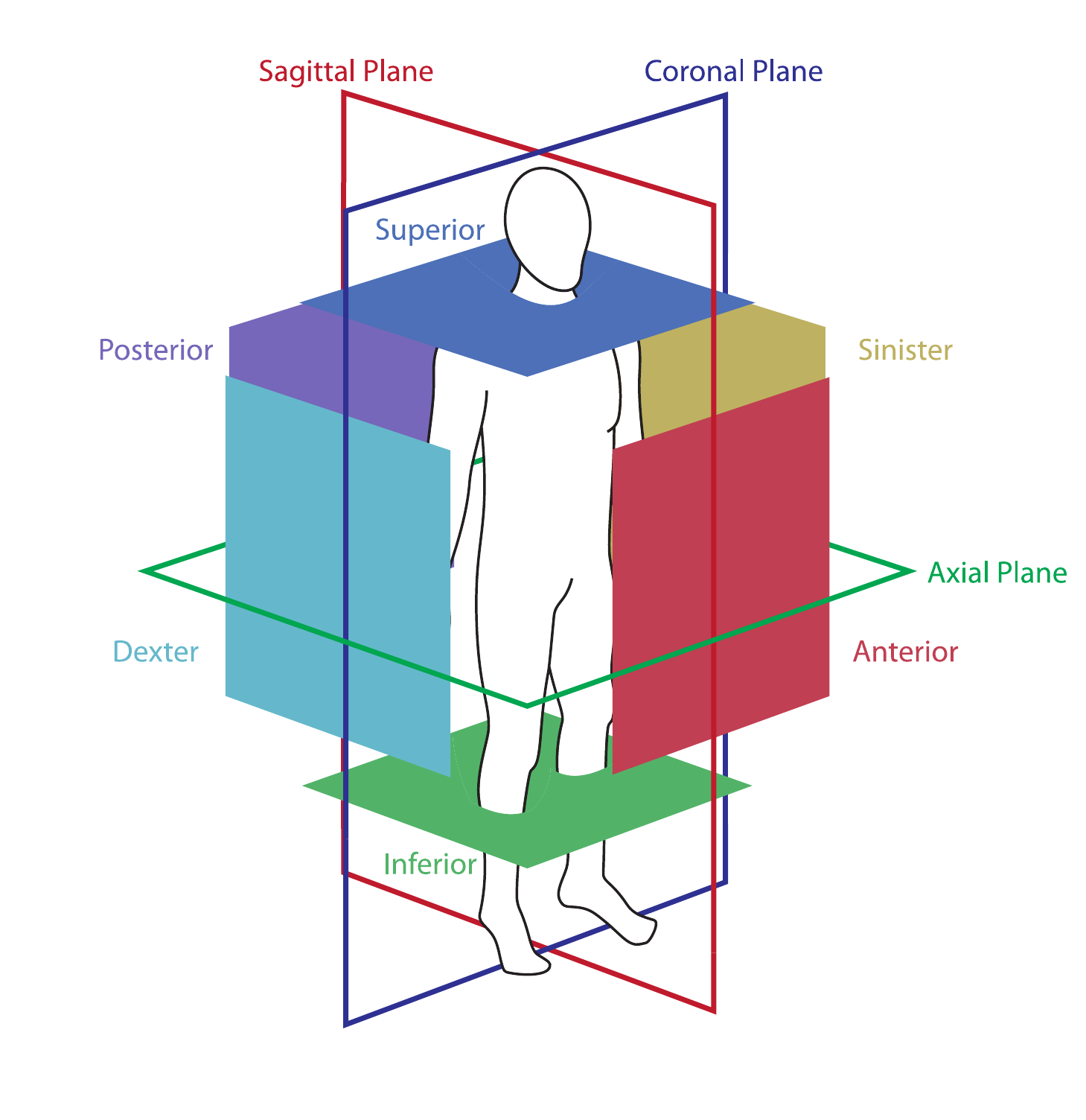}
	\caption{Bounding box walls (opaque squares) and viewing planes (outlined rectangles) shown with anatomical location terminology that is used in the annotation protocol.}
	\label{fig:coords}
\end{figure}

To define a reference, rectangular bounding boxes were manually defined by an expert around anatomical structures in 3D. In chest CT, the anatomical structures of interest were the heart, ascending aorta, descending aorta, and the aortic arch. In cardiac CTA a bounding box around the left ventricle was annotated, and in abdomen CT around the liver. All bounding boxes were tightly drawn around the anatomical structures according to the following protocols (see Figure~\ref{fig:coords} for the used terminology of anatomical positions):

\subsubsection{Heart}
The cardiac bounding box was tightly drawn around the myocardium, excluding epicardial fat. This definition was clear for all bounding box walls, except for the walls in superior anterior direction. In this direction the protocol defined the heart to be contained between the pulmonary artery bifurcation and the cardiac apex.

\subsubsection{Ascending Aorta}
A bounding box demarcating the ascending aorta was mainly defined in axial slices. The inferior-superior boundaries were defined to contain the regions where the ascending aorta had a visible circular shape (occasionally distorted by motion artifacts) up to the boundary of the aortic arch.

\subsubsection{Aortic Arch}
The bounding box delineating the aortic arch was determined mainly in the axial view. The superior wall was defined as the first slice that contained the top of the arch. Thereafter, the inferior wall was defined as the last slice that visualized the ascending aorta connected to the descending aorta. Finally, the middle axial slice was used to outline the other bounding box walls by separating the aortic lumen from other tissue.

\subsubsection{Descending Aorta}
In superior-inferior direction the descending aorta was defined between the aortic arch and the diaphragm. The coronal and sagittal views were used to demarcate the other bounding box walls.

\subsubsection{Left Ventricle}
The left ventricle was defined as the left ventricular blood pool and the surrounding myocardium. Because the myocardium of the left ventricle is connected to other cardiac tissue with similar intensities, positions were estimated for the superior, anterior, and dexter bounding box walls.

\subsubsection{Liver}
Liver tissue has a limited range Hounsfield units and was clearly visible in the contrast enhanced abdomen CT scans. Demarcating the bounding box was straightforward for all walls, except for the sinister wall, because of anatomical variation. The liver sometimes overlapped with the spleen, making a distinction ambiguous in some cases.

Examples of the reference bounding boxes are illustrated in Figure~\ref{fig:data}.

\subsection{Second Observer Annotations}
To compare the performance of the automatic method with the performance of a human, a second expert defined bounding boxes in a subset of 20 scans in each of the three datasets using the same protocol that was used to set the reference.

\begin{figure*}[ht!]
	\centering
	\includegraphics[trim=20 20 20 20, clip, width=.9\textwidth]{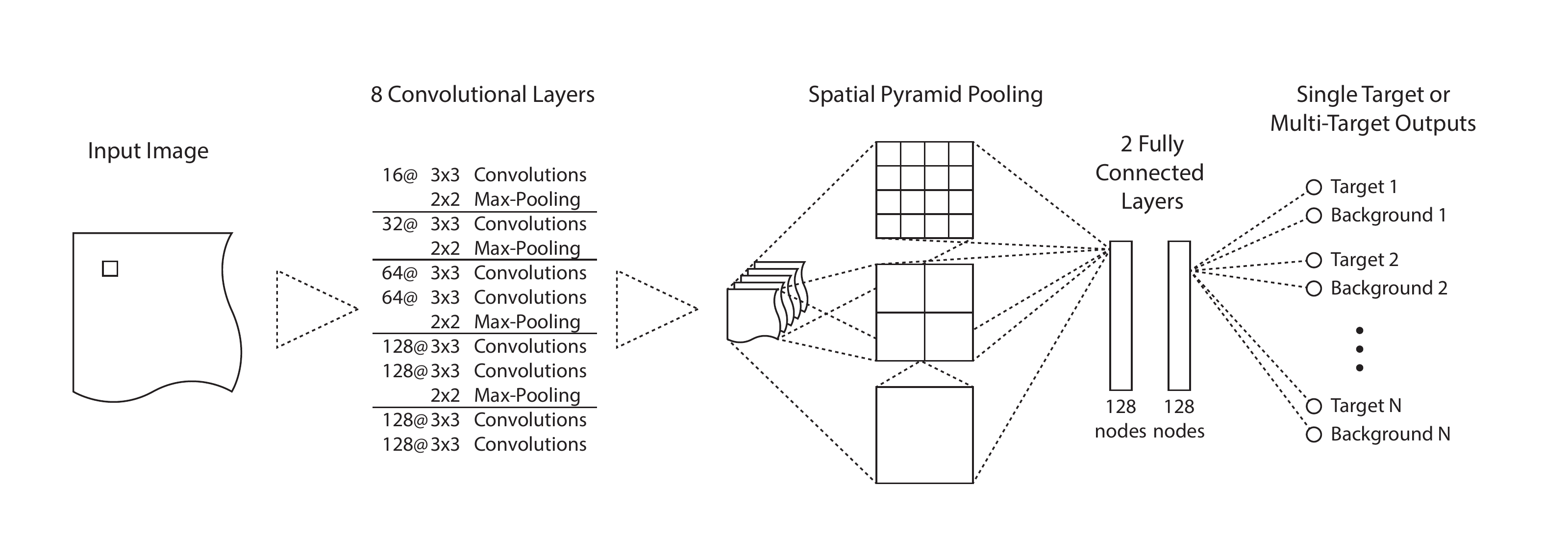}
	\caption{Schematic overview of the bounding box network (BoBNet) architecture. The ConvNet consists of eight convolutional layers of increasing numbers of $3\times3$ kernels: 16, 32, two times 64, and four times 128 kernels. The first, second, fourth, and sixth convolutional layers are downsampled by $2\times2$ max-pooling layers. The eighth convolutional layer is downsampled with a spatial pyramid pooling layer consisting of $4\times4$, $2\times2$, and $1\times1$ max pooling layers. The output feature-maps of the spatial pyramid pooling layer are connected to a sequence of two fully connected layers of 128 nodes each, which are connected to an output layer of $2N$ end-nodes, with $N$ indicating the number of target anatomical structures. Spatial pyramid pooling allows analysis of images of variable input sizes. The $N\times2$ output vector allows to use the softmax function for multi-label classification and therefore multi-organ localization.}
	\label{fig:network}
\end{figure*}

\section{Method}
\label{sec:methods}
The proposed method localizes anatomical structures in 3D medical images by detecting their presence in 2D image slices. The 2D image slices are extracted from the three orthogonal planes of a 3D image: the axial, coronal, and sagittal image planes. Presence of anatomical structures of interest is detected by a single ConvNet, using the 2D image slices as input.

Depending on image acquisition protocol or patient size, medical images may vary in their size. For example, in-plane resolution may depend on patient size and acquisition parameters. To reduce this variation across images and thereby simplify the detection task, all extracted 2D slices are resized to an equal isotropic resolution. This results in 2D slices of substantially varying size.  Note that 2D slices are downsampled and not 3D volumes, because downsampling the full 3D volume would decrease the number of possible image slices that can be extracted. However, standard ConvNets require input images of a fixed size. Hence, input images, i.e. 2D slices extracted from 3D images, could be cropped or padded to a predefined size. Consequently, parts of the anatomical target structures might be cropped out of the image slices, which would hinder detection. To allow a ConvNet to analyze whole input images, regardless of their dimensions, spatial pyramid pooling is used~\cite{he_spatial_2015}. Spatial pyramid pooling forces a fixed length representation of an input image, regardless of its dimensions. In regular pooling, an image (i.e. a feature map) is downscaled based on a fixed kernel size and a fixed kernel stride to obtain a variable output image size. In spatial pyramid pooling, an image is downscaled based on a variable kernel size and a variable kernel stride to force a  fixed output image size. In spatial pyramid pooling several fixed size output images with decreasing dimensions constitute the pyramid, which increases robustness \cite{he_spatial_2015}.

The proposed 3D localization method strongly depends on correct detection of anatomical structures of interest in 2D image slices by a ConvNet. In classification of natural images, AlexNet~\cite{krizhevsky2012}, VGGNet-16~\cite{simonyan2014}, ResNet-34~\cite{heresnet2015} showed excellent classification results. It can be expected that these, or similar networks, can be employed for the detection task in this work. However, these are large networks with many parameters. Given that medical data meet certain acquisition criteria, a simpler architecture with fewer trainable parameters might suffice. Hence, in addition to these large well performing networks, a novel  \textbf{bo}unding \textbf{b}ox \textbf{net}work (BoBNet) is designed. Its architecture is inspired by the designs in~\cite{simonyan2014} and is illustrated in Figure~\ref{fig:network}. BoBNet consists of eight convolutional layers of increasing numbers of $3\times3$ kernels: 16, 32, two times 64, and four times 128 kernels. The first, second, fourth, and sixth convolutional layers are connected to $2\times2$ max-pooling downsampling layers. The eighth convolutional layer is connected to a spatial pyramid pooling layer consisting of $4\times4$, $2\times2$, and $1\times1$ max pooling layers. The output feature-maps of the spatial pyramid pooling layer is connected to a sequence of two fully connected layers of 128 nodes each, which are connected to an output layer of linked $N\times2$ end-nodes, with $N$ indicating the number of target anatomical structures.
	
Rectified linear units are used for activation throughout the network, except in the output layer, which uses the softmax function. The benefit of using a softmax output comes from its relation to the winner-takes-all operation~\cite{bridle1990}. It simplifies classification, because outputs are forced to be either close to 0 or close to 1. This eliminates the need to tune a threshold parameter for the obtained probabilities that indicates presence of the anatomical structures of interest. However, the softmax function is a normalizing transformation, so it cannot handle multi-label classification, i.e. it can not generate probabilities for simultaneously occurring multiple anatomical structures. To enable multi-label outputs, while exploiting the benefits of the softmax function, two linked output-nodes are used per anatomical target structure. One indicates presence of the anatomical target structure and the other indicates its absence.

Due to spatial pyramid pooling, the proposed ConvNet architecture is able to analyze input images of varying size. During training however, mini-batches of slices extracted from different scans and different orientations, thus images of different sizes, are presented to the network. Furthermore, current ConvNet implementations~\cite{jia2014caffe,torch,2016arXiv160502688short,tensorflow2015}, only allow training of ConvNets with tensors of a fixed size in each dimension. To overcome this issue, like in \cite{he_spatial_2015}, a ConvNet is trained with several predefined sizes of the input, by padding or cropping input images when necessary. But unlike in \cite{he_spatial_2015}, where input sizes alternated per epoch, input image sizes vary per mini-batch. 
So, in each minibatch input images are randomly cropped or padded based on the first and third quartiles of image dimensions of the images in the given minibatch. This resulted in four different combinations of input image sizes. The minimum input size is set to $224\times224$ pixels and cropping or padding is constrained such that most of the original image is retained. In addition, image slices are randomly rotated by a maximum of $10^\circ$ in clockwise or anti-clockwise direction. After training the ConvNet is able to generate posterior probabilities indicating presence of anatomical structures of interest from original full input image slices.

\begin{figure}
	\centering
	\includegraphics[width=\columnwidth]{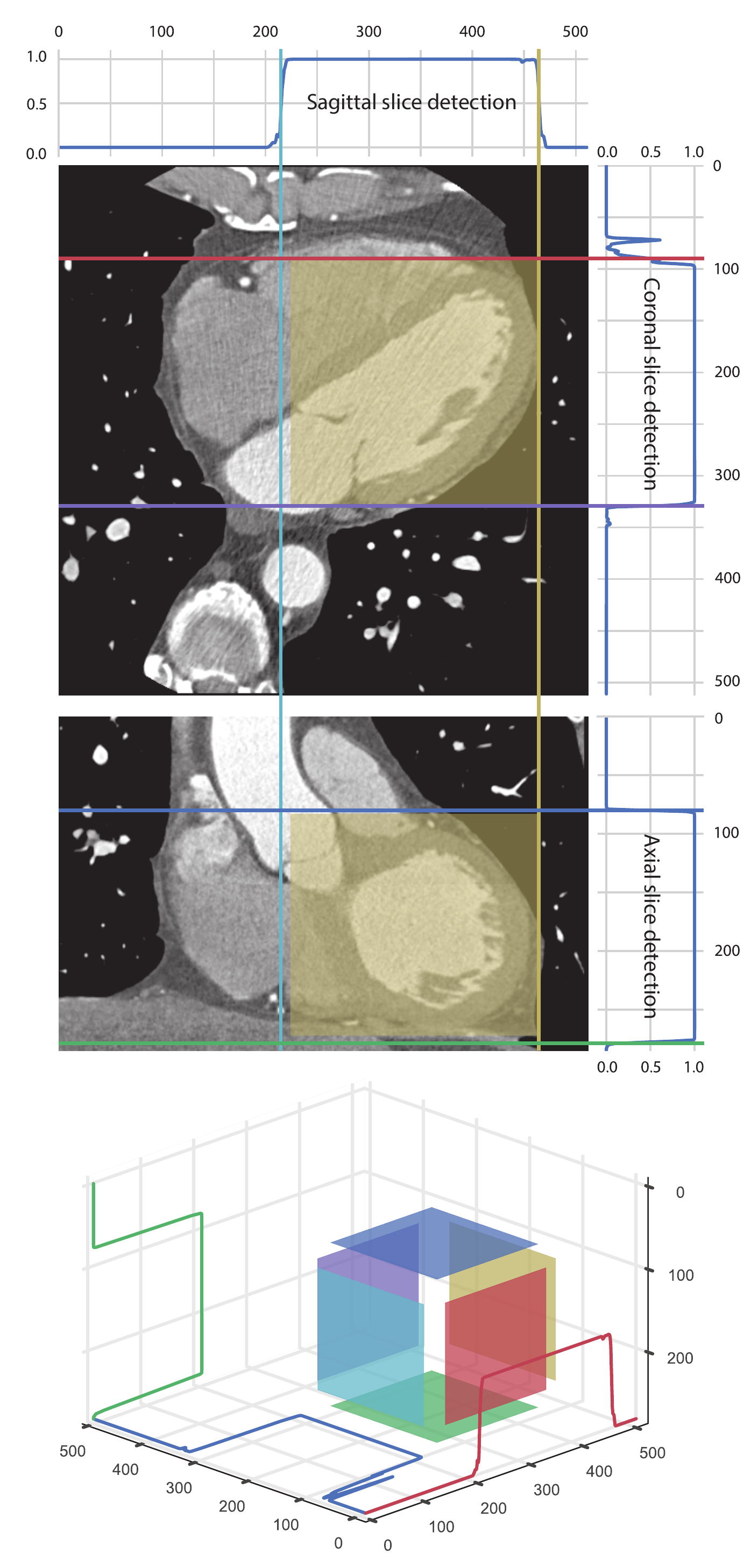}
	\caption{Example of localization of the left ventricle in a 3D cardiac CTA image by detecting its presence in 2D image slices extracted from the three orthogonal image planes. The plots show automatic detection results for image slices extracted from the sagittal, coronal, and axial planes. Axial (top)  and coronal (middle) image slices indicate the reference bounding box and the predicted bounding box walls. The predicted borders are shown in the colors used in Figure~\ref{fig:coords}. The bottom figure illustrates combination of the detection results to localize the anatomical target structure by forming a 3D bounding box.}
	\label{fig:responses}
\end{figure}

Thereafter, to localize the anatomical target structures, outputs of the ConvNet for axial, coronal, and sagittal slices are combined. A 3D bounding box ($B$), i.e. a rectangular mask, is created as follows:
\begin{align*}
B_{i,j,k}=
\begin{cases} 
1 	 & \text{if }\textbf{s}_i\geq t \wedge \textbf{c}_j\geq t \wedge \textbf{a}_k\geq t \\
0       & \text{elsewhere }
\end{cases}
\end{align*}
where $B$ is the bounding box, and $t$ is a threshold on the posterior probabilities detecting presence of the anatomical structure of interest indicated by the vectors $\textbf{s}$, $\textbf{c}$, and $\textbf{a}$, which are respectively the sagittal, coronal, and axial slices. The indices $i$, $j$, and $k$ indicate slice positions and thus indicate image coordinates in the CT volume. The threshold $t$ is set to $0.5$. To ensure a single bounding box, only the largest 3D connected component is retained.

\section{Evaluation}
To evaluate \textit{detection} performance of ConvNet architectures, the F1 score was used:
\begin{equation*}
F1 = 2\cdot \frac{precision\cdot recall}{precision + recall}  
\end{equation*}	
To evaluate whether there is a significant difference in detection performance between architectures, McNemar's test was used~\cite{salzberg1997}.

To evaluate \textit{localization} performance of the employed method, distances from the automatically determined bounding boxes to the reference bounding boxes were determined. First, distances between automatic bounding box walls and the reference bounding box walls were computed. In addition, the distance between automatic and reference bounding box centroids was computed. The bounding boxes of the second expert and the reference bounding boxes were compared using the same metric.

\section{Experiments and Results}
\label{sec:experiments}
\subsection{Experimental Details}
Bounding boxes were created in chest CT, cardiac CTA, and abdomen CT. Experiments were separately performed for each set. The images were resampled to a fixed resolution: 1.5\,mm per pixel for chest CT and abdomen CT, and 1.0\,mm per pixel for cardiac CTA.
For each set, 50\% of the randomly selected scans were assigned to the training set and the remaining 50\% of the scans were assigned to the test set. From the training set, 10\% of the scans were used for validation. The test set was not used during method development in any way.
 
ConvNet weights were initialized with a Glorot-uniform distribution~\cite{glorot2010} and trained in 30 epochs by minimizing cross-entropy with Nesterov's accelerated gradient descent~\cite{nesterov}. The initial learning rate was set at 0.01 and it was reduced by a factor of 10 every 10 epochs. The ConvNet parameters were L2-regularized with a weight of $5\cdot10^{-4}$. Dropout of 50\% was applied to the nodes of the fully connected hidden layers. Image slices were provided once per epoch to the ConvNet in randomized order in mini-batches of 64 image slices.

All experiments were performed using Theano~\cite{2016arXiv160502688short} and Lasagne~\cite{lasagne}.

\subsection{2D Detection: Comparison of ConvNet Architectures}
\begin{figure}
	\centering
	\includegraphics[trim=0 0 0 0, clip, width=.45\textwidth]{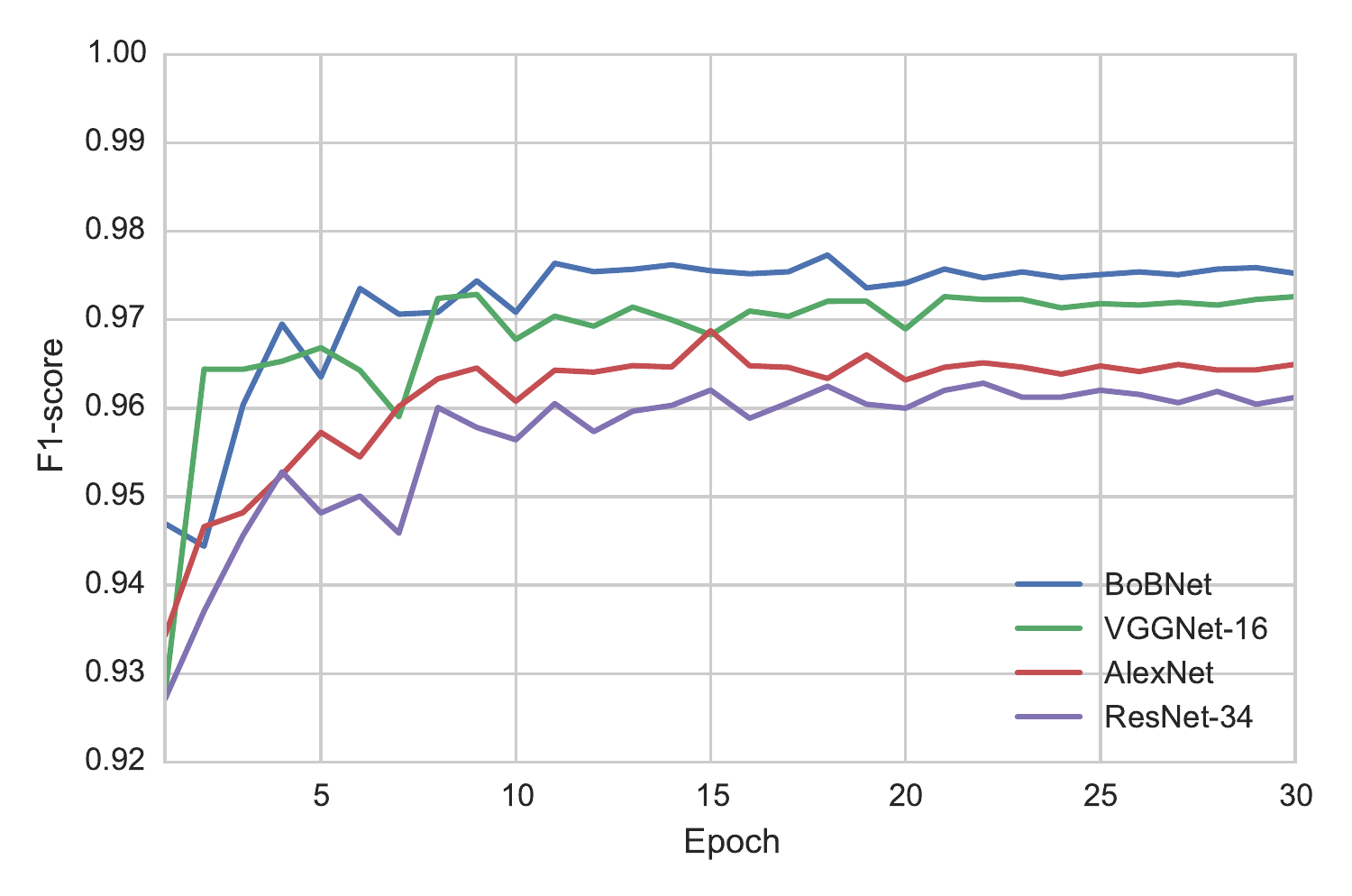}
	\caption{F1-scores of four evaluated ConvNet architectures showing detection performance of the left ventricle in cardiac CTA image slices. The F1-score was determined in every epoch on the validation set. For each ConvNet plateaus formed after 15 epochs of training. This indicated that no overfitting occurred.}
	\label{fig:f1scores}
\end{figure}

\begin{table}[b]
	\centering
	\caption{F1-scores achieved by the ConvNets evaluated on left ventricle detection in image slices from cardiac CTA.}
	\label{tab:f1scores}
	\begin{tabular}{ll}
		ConvNet Architecture & F1-score\\
		\hline
		BoBNet & 0.967 \\
		VGGNet-16 \cite{simonyan2014}           & 0.963 \\
		ResNet-34 \cite{heresnet2015}         & 0.960 \\
		AlexNet \cite{krizhevsky2012}          & 0.959 \\
	\end{tabular}
\end{table}

\begin{table}
	\centering
	\caption{McNemar tests between different ConvNet architectures evaluated on detection of the left ventricle in image slices from cardiac CTA. Significant $p$-values ($<0.01$) are shown between all ConvNets, except ResNet-34 and AlexNet.}
	\label{tab:mcnemar}
	\begin{tabular}{l|rrrr}
		&BoBNet&VGGNet-16& ResNet-34  & AlexNet    \\
		\hline
		BoBNet        & & $\ll0.01$& $\ll0.01$& $\ll0.01$\\
		VGGNet-16      & $\ll0.01$ & & $\ll0.01$& $\ll0.01$\\
		ResNet-34      & $\ll0.01$  &$\ll0.01$ & & $\phantom{\ll}0.06$\\
		AlexNet	& $\ll0.01$& $\ll0.01$& $0.06$&\\
	\end{tabular}
\end{table}

The proposed 3D localization method strongly depends on correct detection of anatomical structures of interest in 2D image slices by a ConvNet (see Figure~\ref{fig:responses}). Hence, the performances of AlexNet~\cite{krizhevsky2012}, VGGNet-16~\cite{simonyan2014}, ResNet-34~\cite{heresnet2015} and BoBNet were evaluated on detection of the left ventricle in cardiac CTA. In AlexNet, VGGNet-16, and ResNet-34 the final pooling layer was replaced by the spatial pyramid pooling layer (described in Section~\ref{sec:methods}) to allow a fair comparison of ConvNet architectures.

\begin{figure*}[t]
	\centering
	\begin{subfigure}[b]{.9\columnwidth}
		\includegraphics[trim=0 0 0 0, clip, width=\columnwidth]{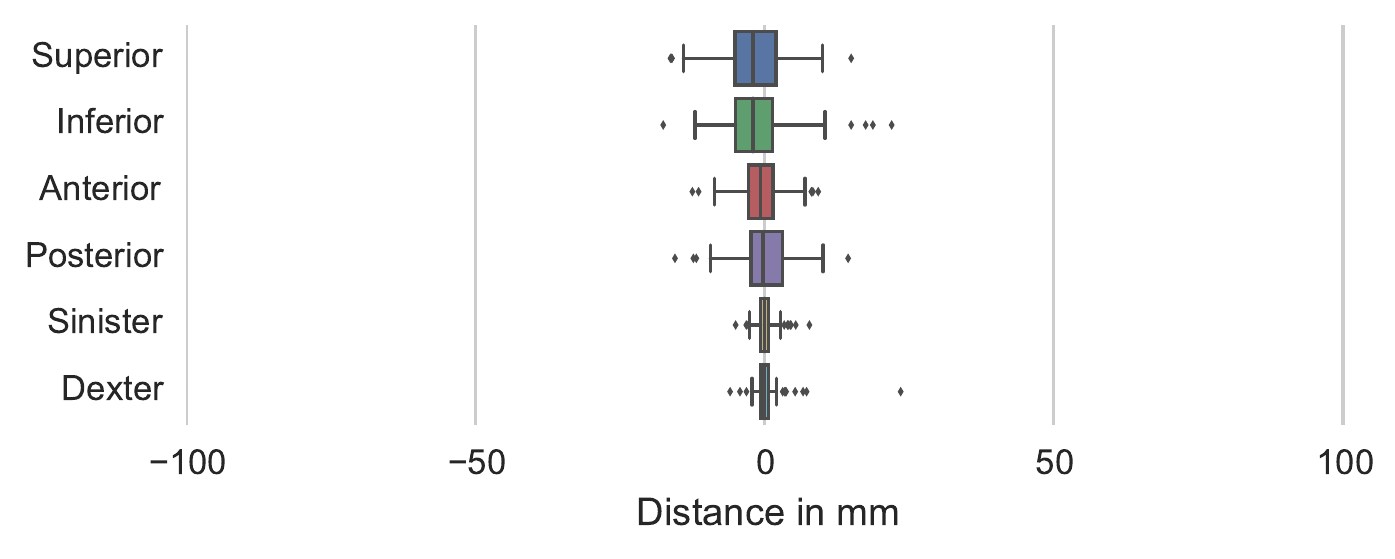}
		\caption{Heart (Chest CT)}
	\end{subfigure}
	\vspace{.5cm}
	\begin{subfigure}[b]{.9\columnwidth}
		\includegraphics[trim=0 0 0 0, clip, width=\columnwidth]{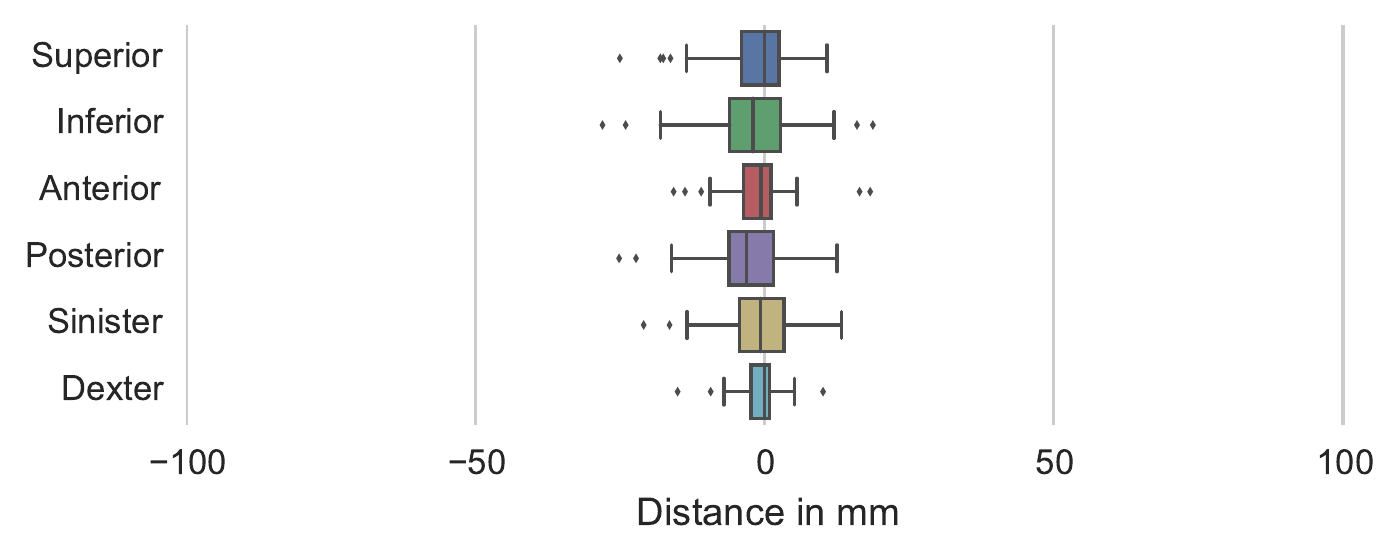}
		\caption{Ascending Aorta (Chest CT)}
	\end{subfigure}\\
	\vspace{.5cm}
	\begin{subfigure}[b]{.9\columnwidth}
		\includegraphics[trim=0 0 0 0, clip, width=\columnwidth]{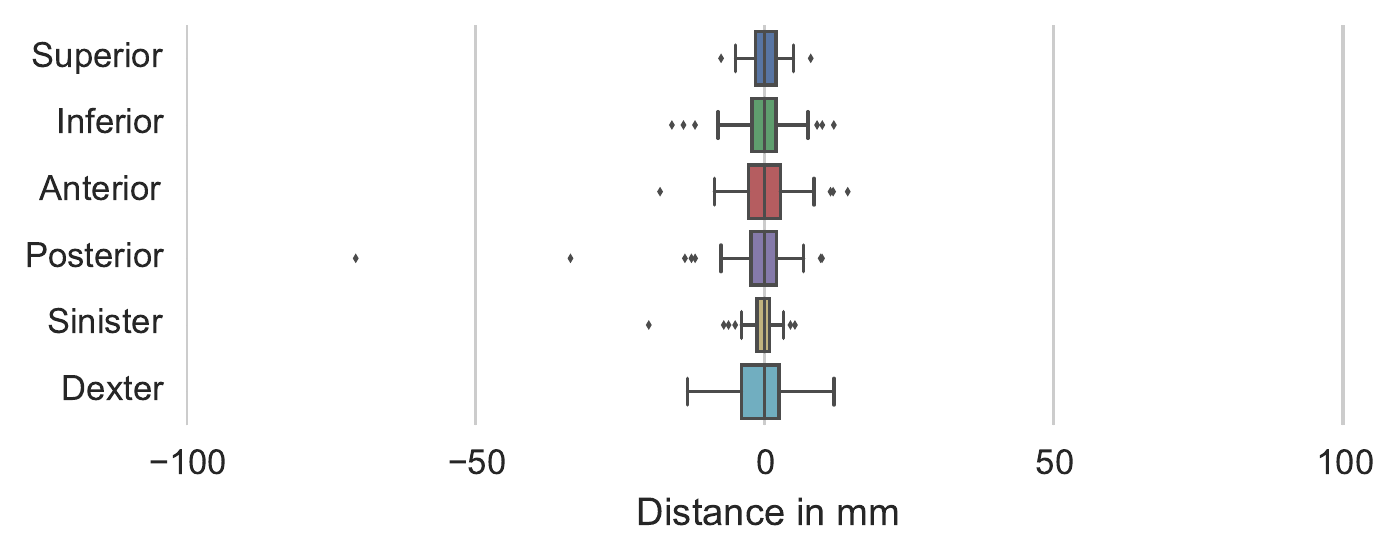}
		\caption{Aortic Arch (Chest CT)}
	\end{subfigure}
	\vspace{.5cm}
	\begin{subfigure}[b]{.9\columnwidth}
		\includegraphics[trim=0 0 0 0, clip, width=\columnwidth]{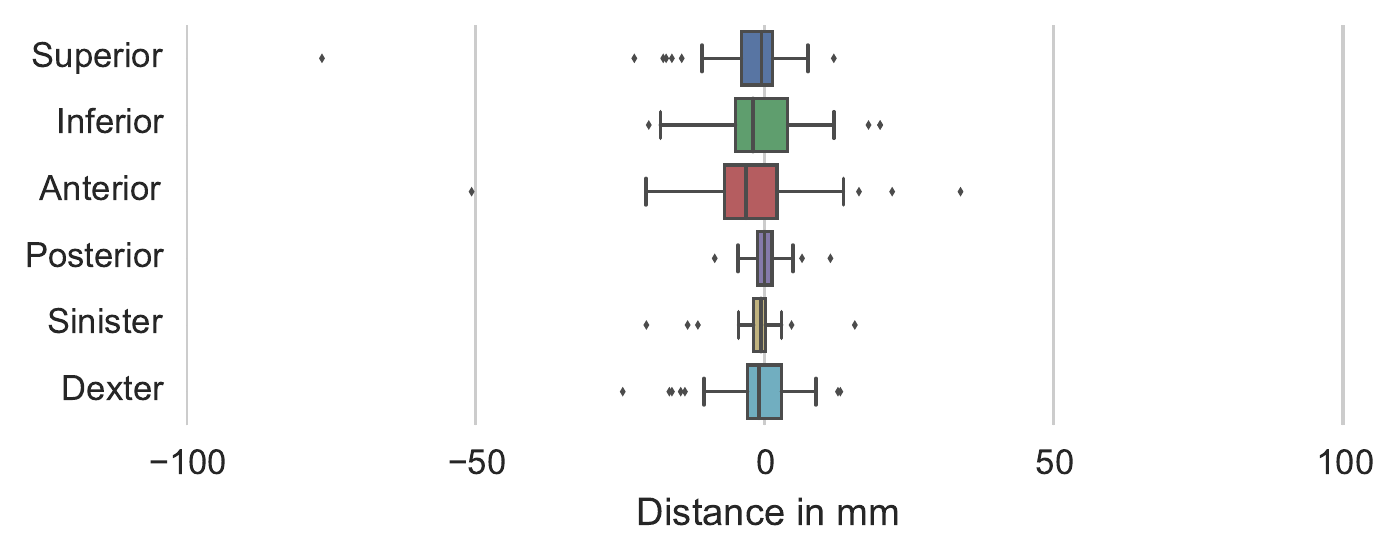}
		\caption{Descending Aorta (Chest CT)}
	\end{subfigure}\\
	\vspace{.5cm}
	\begin{subfigure}[b]{.9\columnwidth}
		\includegraphics[trim=0 0 0 0, clip, width=\columnwidth]{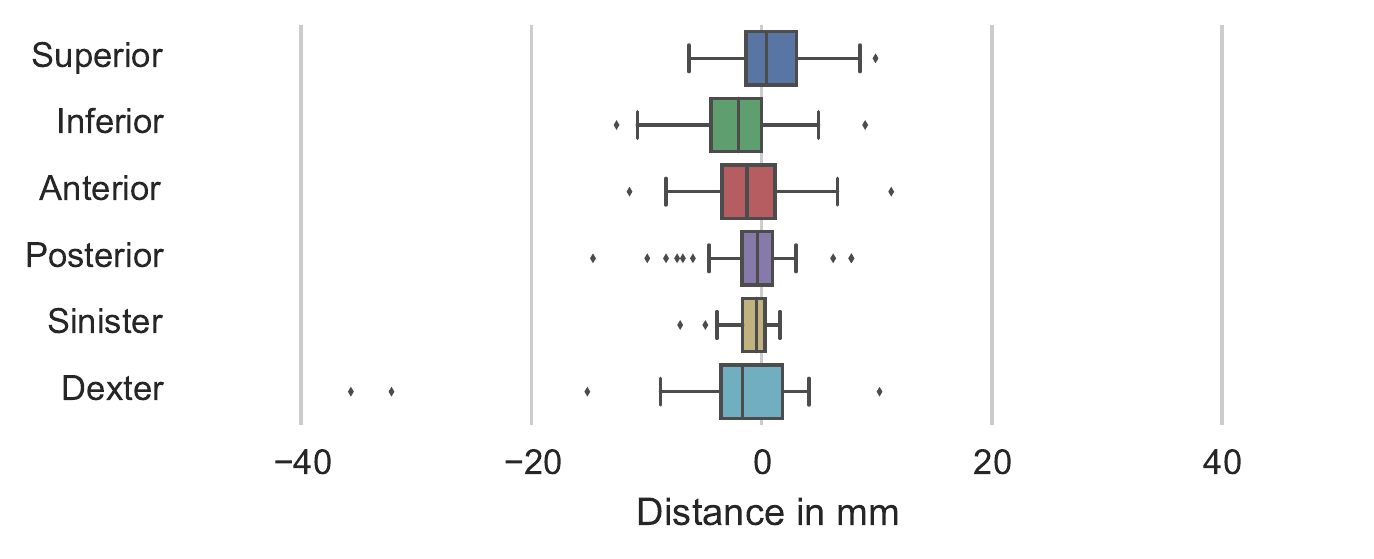}
		\caption{Left Ventricle (CCTA)}
	\end{subfigure}
	\vspace{.5cm}
	\begin{subfigure}[b]{.9\columnwidth}
		\includegraphics[trim=0 0 0 0, clip, width=\columnwidth]{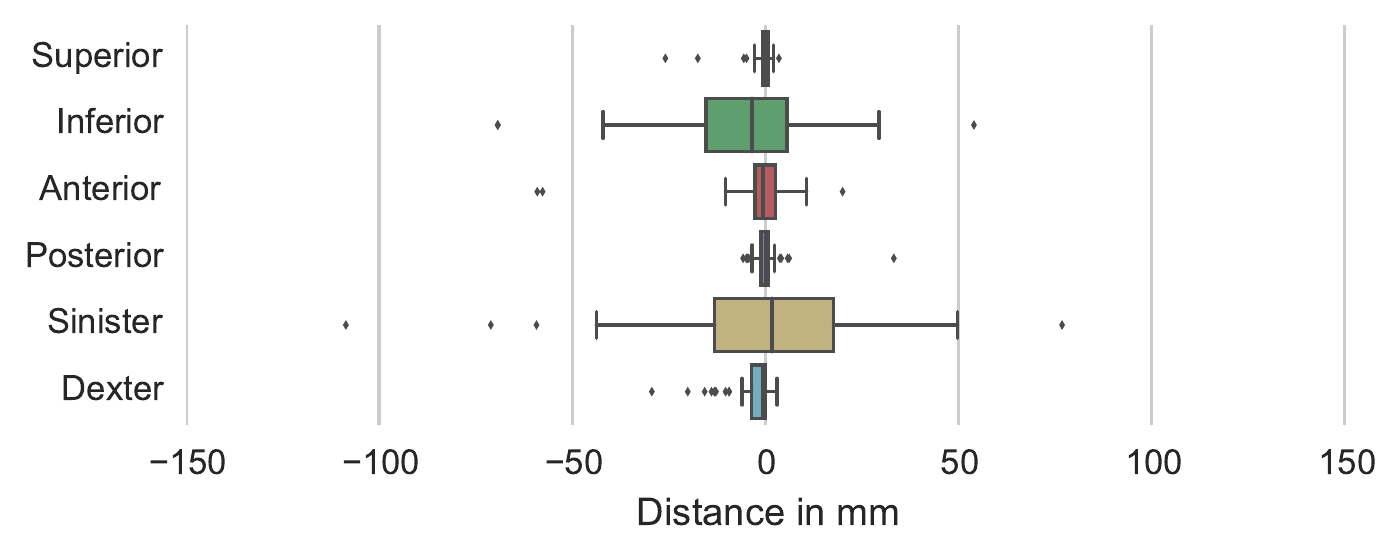}
		\caption{Liver (Abdomen CT)}
	\end{subfigure}
	\caption{Boxplots showing results of single anatomy localization of the heart (a), ascending aorta (b), aortic arch (c), descending aorta (d) in chest CT; left ventricle (e) in cardiac CTA; and liver (f) in abdomen CT. Distances between automatic and reference bounding box walls are shown. Negative numbers correspond to undersegmentations and positive numbers correspond to oversegmentations. Refer to Figure~\ref{fig:coords} for the positions of the bounding box walls. Note that every dataset has a different scale of the horizontal axis.}
	\label{fig:distances}
\end{figure*}

To ensure there was no overfitting, detection performance was monitored during training on the validation set with the F1-score. Figure~\ref{fig:f1scores} shows that all ConvNets converged after 15 epochs without an indication of overfitting. Thus, ConvNets trained after 30 epochs were used for evaluation. Table~\ref{tab:f1scores} lists the obtained F1 scores. The best results were achieved using BoBNet. Table~\ref{tab:mcnemar} lists $p$-values of McNemar tests comparing detection results between ConvNets. The results show that there was a significant difference ($p<0.01$) between all compared ConvNets except between AlexNet and ResNet-34. Given that the proposed BoBNet achieved the best performance, it was used in all subsequent experiments.

\begin{table}
	\centering
	\caption{Localization of single anatomical structures. Mean and standard deviation (in mm) of the wall distances and centroid distances between the automatically determined and the reference bounding boxes are listed for the localization of the heart, ascending aorta (AAo), aortic arch (AoArch), and descending aorta (DAo) in chest CT, left ventricle (LV) in cardiac CTA, and liver in abdomen CT.}
	\label{tab:singleanatomy}
	\begin{tabular}{llll}
		Dataset                   & Structure      & Wall dist. (mm)  & Centroid dist. (mm)          \\
		\hline
		\multirow{4}{*}{Chest CT} & Heart        & $\phantom{0}3.11 \pm \phantom{0}3.43$ & $\phantom{0}5.01 \pm \phantom{0}3.30$ \\
		& AAo            & $\phantom{0}4.32 \pm \phantom{0}4.16$ & $\phantom{0}6.15 \pm \phantom{0}2.88$ \\
		& AoArch         & $\phantom{0}2.93 \pm \phantom{0}4.18$ & $\phantom{0}4.32 \pm \phantom{0}4.10$ \\
		& DAo            & $\phantom{0}4.09 \pm \phantom{0}5.61$ & $\phantom{0}6.93 \pm \phantom{0}5.30$ \\ 
		Cardiac CTA               & LV & $\phantom{0}2.97 \pm \phantom{0}3.67$ & $\phantom{0}4.47 \pm \phantom{0}3.38$ \\
		Abdomen CT                & Liver          & $\phantom{0}8.87 \pm 15.00$ & $16.93 \pm 11.54$ \\
	\end{tabular}
\end{table}

\subsection{3D Localization of Single Anatomical Structures}
\begin{figure*}[ht!]
	\flushleft
	\begin{subfigure}[]{1\textwidth}
		\includegraphics[trim=0 0 0 0, clip, width=\textwidth]{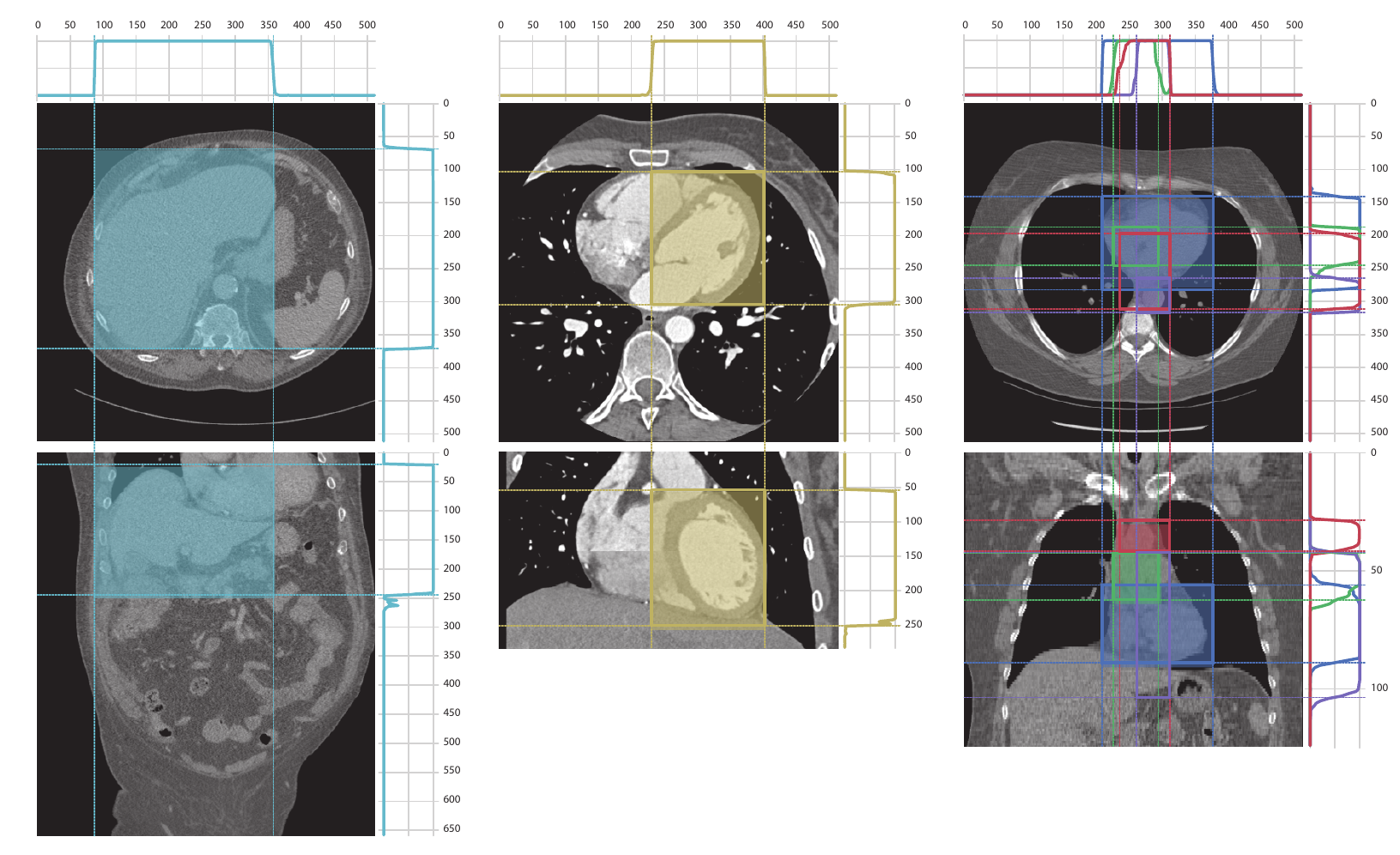}
	\end{subfigure}
	
	\begin{subfigure}[]{1\textwidth}
		\includegraphics[trim=0 0 0 0, clip, width=\textwidth]{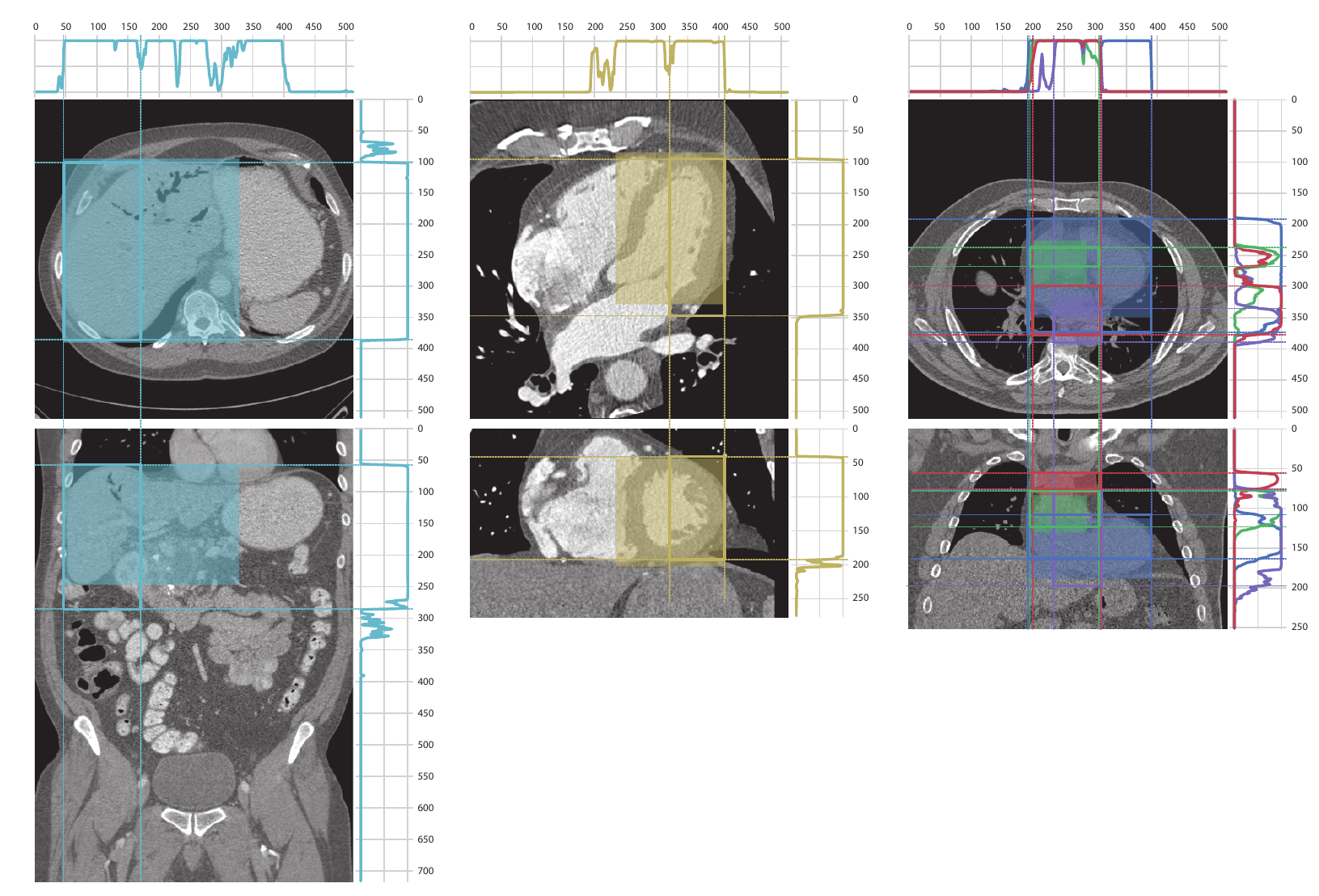}
	\end{subfigure}\\
	{\vspace{-3.6cm}\hspace{6.5cm}}
	\captionsetup{width=.6\textwidth,justification=justified,slc=off}
	\captionbox{The best (top) and the worst (bottom) localization results of a single anatomical structure (the liver in abdomen CT on the left and the left ventricle in cardiac CTA in the middle) and localization of multiple anatomical structures (heart, ascending aorta, aortic arch, and descending aorta in chest CT on the right). Reference annotations are shown as opaque areas in colors as described in Figure~\ref{fig:data} and predicted bounding boxes are shown by solid lines. The best localization results show smooth posterior probabilities over all slices, while the worst results show unstable posterior probabilities over slices influencing the final localization.\label{fig:goodbad}}
	
\end{figure*}
Localization of single anatomical structures was evaluated for all anatomical target structures in all three datasets: the heart, ascending aorta, aortic arch, and descending aorta in chest CT scans; the left ventricle in cardiac CTA; and the liver in abdomen CT. One abdomen CT was removed from analysis, because of a large internal hemorrhage, making it hard to distinguish anatomy in the abdomen of this specific patient. Localization did not fail in any of the test scans.
\begin{figure*}[ht!]
	\centering
	\begin{subfigure}[b]{.9\columnwidth}
		\includegraphics[trim=0 0 0 0, clip, width=\columnwidth]{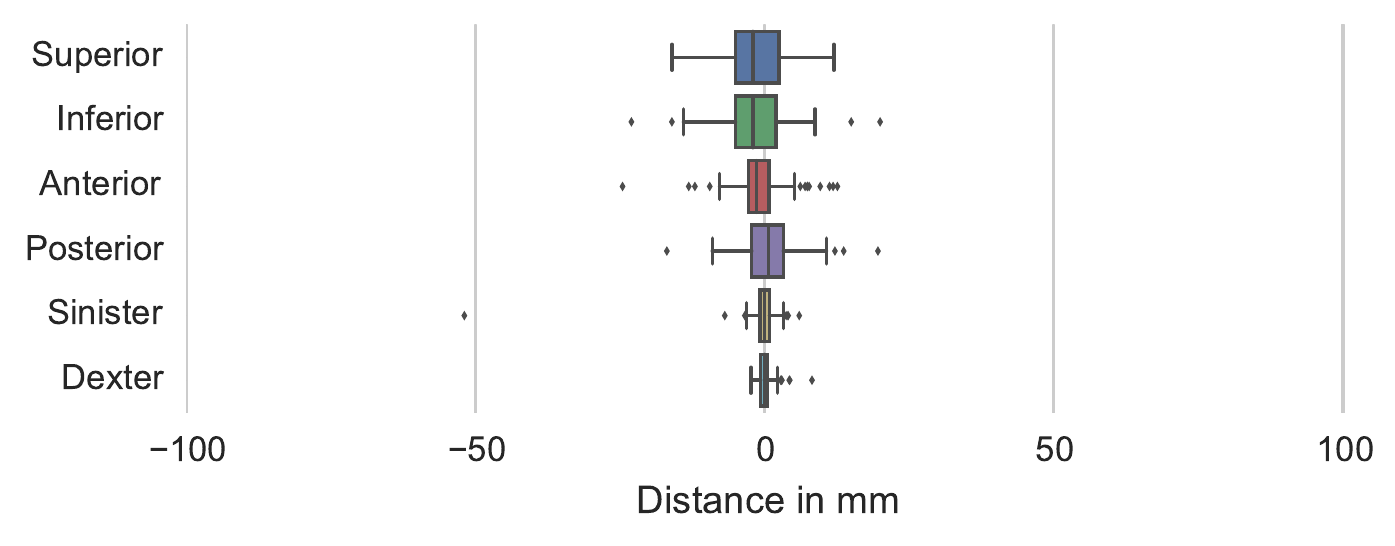}
		\caption{Heart (Chest CT)}
	\end{subfigure}
	\vspace{.5cm}
	\begin{subfigure}[b]{.9\columnwidth}
		\includegraphics[trim=0 0 0 0, clip, width=\columnwidth]{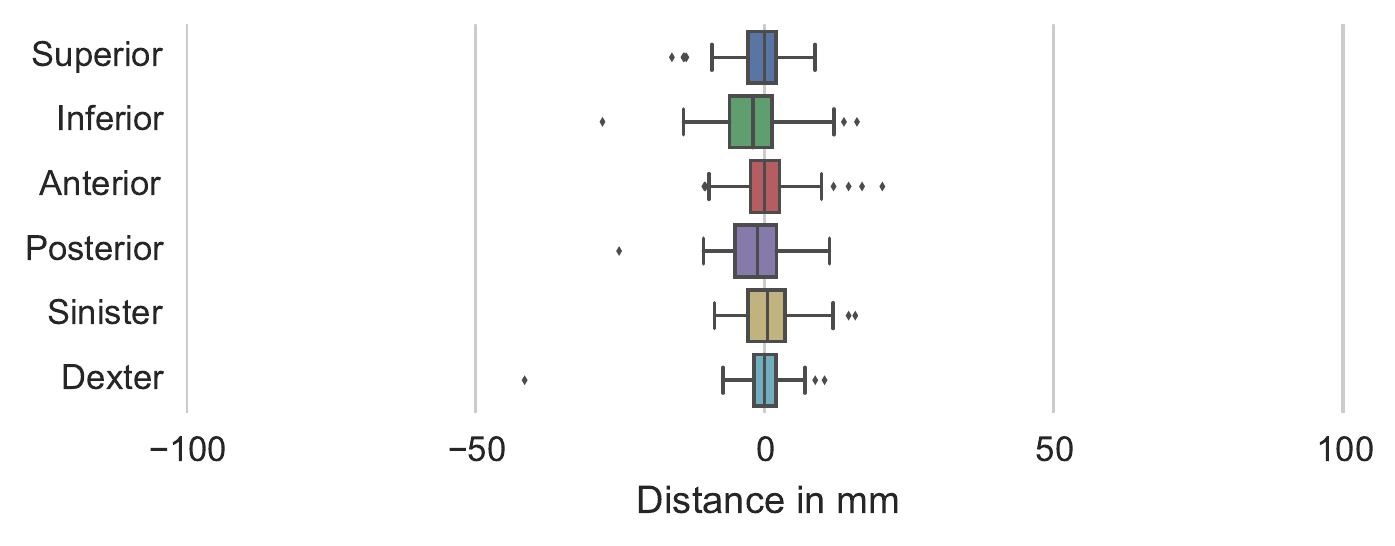}
		\caption{Ascending Aorta (Chest CT)}
	\end{subfigure}\\
	\vspace{.5cm}
	\begin{subfigure}[b]{.9\columnwidth}
		\includegraphics[trim=0 0 0 0, clip, width=\columnwidth]{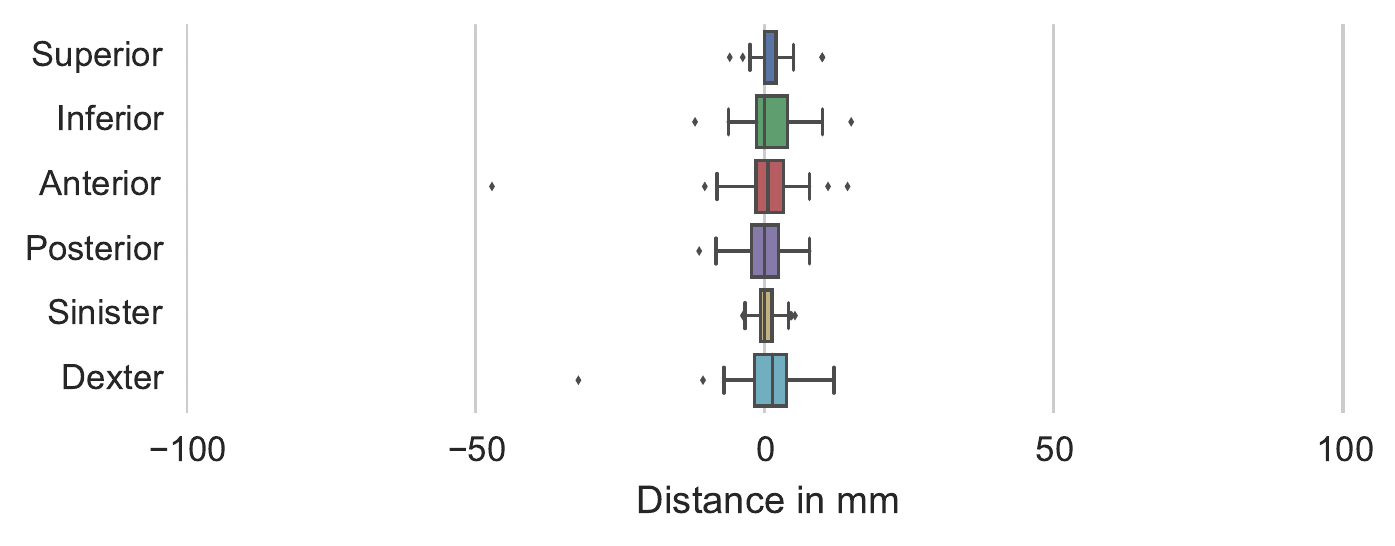}
		\caption{Aortic Arch (Chest CT)}
	\end{subfigure}
	\vspace{.5cm}
	\begin{subfigure}[b]{.9\columnwidth}
		\includegraphics[trim=0 0 0 0, clip, width=\columnwidth]{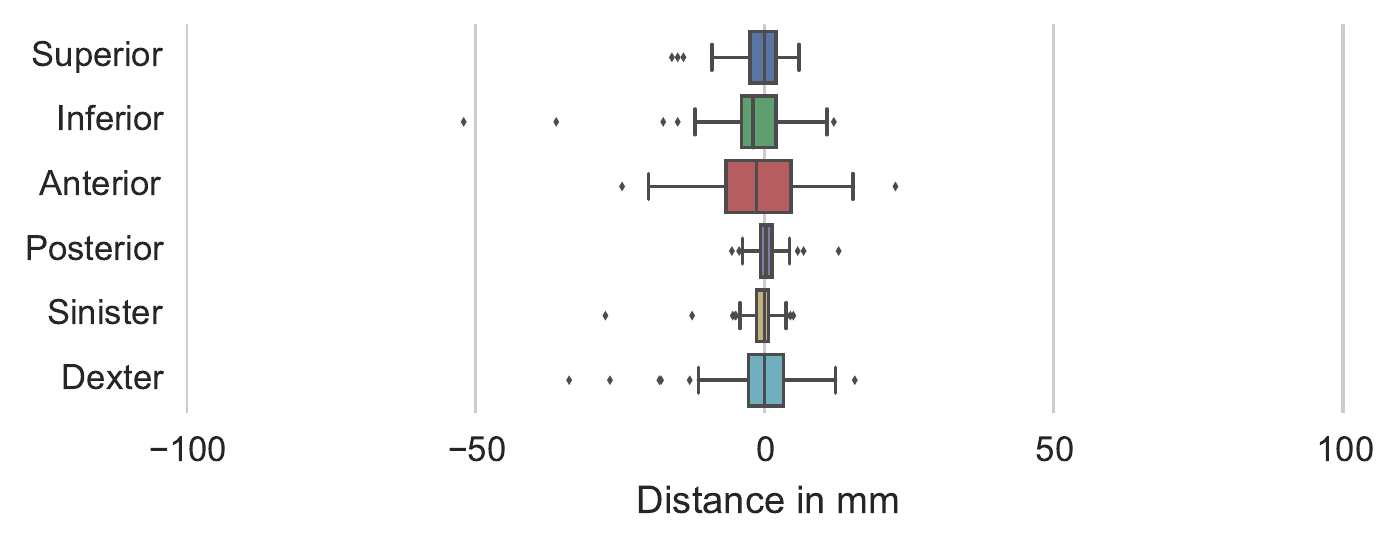}
		\caption{Descending Aorta (Chest CT)}
	\end{subfigure}\\
	
	\caption{Boxplots showing results of multiple anatomy localization in chest CT scans. Distances between automatic and reference bounding boxes walls are shown. Negative numbers correspond to undersegmentations and positive numbers correspond to oversegmentations. Refer to Figure~\ref{fig:coords} for the positions of the bounding box walls.}
	\label{fig:distances_multilabel}
\end{figure*}

Table~\ref{tab:singleanatomy} lists the mean and standard deviation of wall distances and centroid distances between the automatically obtained and reference bounding boxes for all scans in the test set. The results show that the method achieved the smallest error for localization of the aortic arch, and the largest error for localization of the liver.

\begin{table}
	\centering
	\caption{Localization of multiple anatomical structures. Mean and standard deviation (in mm) of the wall distances and centroid distances between the automatically determined and the reference bounding boxes are listed for the localization of the heart, ascending aorta (AAo), aortic arch (AoArch), and descending aorta (DAo) in chest CT.}
	\label{tab:multianatomy}
	\begin{tabular}{llll}
		Dataset                   & Structure      & Wall dist. (mm)  & Centroid dist. (mm)          \\
		\hline
		\multirow{4}{*}{Chest CT} &Heart  &$3.24 \pm 4.03$ & $5.25 \pm 3.96$ \\
		& AAo&$3.92 \pm 3.88$ & $5.54 \pm 3.20$ \\
		& AoArch&$2.74 \pm 3.24$ & $4.10 \pm 3.11$ \\
		& DAo&$3.76 \pm 4.78$ & $6.24 \pm 4.17$ \\
		\hline
		& Total &$3.42 \pm 4.05$&$5.28 \pm 3.72$\\
	\end{tabular} 
\end{table}

Figure~\ref{fig:distances} shows boxplots for the separate walls of the bounding boxes revealing differences in performance among different bounding box walls. Similar results were obtained in localization of most bounding box walls, but several larger outliers occurred in e.g. localization of the posterior wall of the aortic arch bounding box, and of the superior and anterior wall of the descending aorta bounding box. Left ventricle localization shows very consistent results without large outliers. In liver localization large variation in localization of the sinister and inferior bounding box walls is visible. Figure~\ref{fig:goodbad} illustrates the best and the worst segmentation results for the liver (left) and for the left ventricle (middle).

\subsection{3D Localization of Multiple Anatomical Structures}
Simultaneous localization of multiple anatomical structures was evaluated in chest CT scans. Bounding boxes were obtained around four anatomical target structures: the heart, ascending aorta, aortic arch, and descending aorta. Localization did not fail in any test scan. Table~\ref{tab:multianatomy} lists the mean and standard deviation of absolute wall distances and centroid distances between the automatically obtained and reference bounding boxes. Figure~\ref{fig:distances_multilabel} shows boxplots for the separate wall distances of the bounding boxes. Better overall results were obtained compared to single anatomy localization with less severe outliers. Figure~\ref{fig:goodbad} (right) illustrates the best and the worst segmentation results for the localization of multiple structures in chest CT.
\begin{table}
	\centering
	\caption{Localization of the second observer. Mean and standard deviation (in mm) of the wall distances and centroid distances between the second observer and the reference bounding boxes are listed for the localization of the heart, ascending aorta (AAo), aortic arch (AoArch), and descending aorta (DAo) in chest CT, left ventricle (LV) in cardiac CTA, and liver in the abdomen CT. Note that the comparison was performed in a subset of 20 test scans.}
	\label{tab:secondObserver}
	\begin{tabular}{llll}
		Dataset                   & Structure      & Wall dist. (mm)  & Centroid dist. (mm)          \\
		\hline
		\multirow{4}{*}{Chest CT} & Heart       & $1.37 \pm 1.46$    & $1.72 \pm 0.93$ \\
		& AAo         & $1.66 \pm 1.59$   & $2.49 \pm 1.40$  \\
		& AoArch       & $1.99 \pm 1.64$  & $3.02 \pm 1.22$  \\
		& DAo           & $2.04 \pm 2.76$  & $3.45 \pm 2.85$ \\
		Cardiac CTA               & LV & $1.29 \pm 1.30$ & $1.84 \pm 0.71$  \\
		Abdomen CT                & Liver         & $3.21 \pm 6.73$ & $5.15 \pm 6.89$ 
	\end{tabular}
\end{table}

\subsection{Localization Time}
Localization speed is mainly limited by the throughput of anatomical slices through a ConvNet. Detecting one or multiple anatomical target structures, on the other hand, did not influence speed. Detection in a single image slice (of $256\times256$ pixels) took about 1.2--1.7\,ms on a state-of-the-art GPU using BoBNet. Assuming that anatomy detection would take 1.5\,ms per image slice, localization of a structure in a medical image that consists of $512\times512\times400$ voxels (i.e. 1,424 image slices) would take about 2.1\,s.

\subsection{Comparison with a Second Observer}
Localizations of the second expert annotations were compared with the reference. This was evaluated using a randomly selected subset of 20 test scans from each data set. Table~\ref{tab:secondObserver} lists the obtained results.

\begin{table*}
	\center
	\caption{Results reported in previous work (top rows), our preliminary work (middle rows), and our current method (bottom rows). The table lists the modality used and the anatomical target structure(s), the number of scans in the training and test sets, mean and standard deviations for wall and centroid distance between automatic and reference bounding box walls, and approximate processing times. Note that only a direct comparison was possible between our preliminary and current work, because only these were evaluated using the same images and same anatomical targets. Localization results of anatomical target structures evaluated in this work are shown separately.}
	\label{tab:comparison_distance}
	\begin{tabular}{lllccccc}
		Method & Modality & Target & Training & Test &  Wall dist. (mm) & Centroid dist. (mm) & Time in s ($\approx$)\\
		\hline
		Zheng et al.\cite{zheng2009} & 2D MRI & Left ventricle & 400 & 395 & -- &$13.5\pm\textrm{--}\phantom{1.5}$&0.5\\
		Ghesu et al.\cite{ghesu2016} & 3D Ultrasound & Aortic valve &2,481& 410 & -- & $1.8\pm1.3$ & 0.5\\
		Criminisi et al.\cite{criminisi2013} & CT & 26 structures &318& 82&  $13.3\pm13.0$& -- & 4\\
		& & ~~~\textit{heart}& & &$13.4\pm10.5$ & & \\
		& & ~~~\textit{left ventricle}& & & $14.1\pm12.3$& & \\
		& & ~~~\textit{liver}& & &$15.7\pm14.5$ & & \\
		Gauriau et al.\cite{gauriau2015} & CT & 6 structures & 50 & 80 & $8.8\pm5\phantom{.8}$ & -- & 5.7\\
		& & ~~~\textit{liver}& & &$10.7\pm4\phantom{1.0}$ & & \\
		Lu et al.\cite{lu2016} & CT & Right kidney & 450 & 49 &--& $7.8\pm9.4$ & -- \\
		\hline
		\multirow{3}{2.2cm}{De Vos et al.\cite{devos2016}} 
		
		& Cardiac CTA & Left ventricle & 50 & 50 & $3.2\pm3.4$ & $4.9\pm2.8$ & $2.6$\\
		& Abdomen CT & Liver & 50 & 49 & $10.8\pm15.8$ & $19.0\pm10.5$ & $7.0$\\
		& Chest CT & 4 structures & 100 & 100 & $3.4\pm3.7$ & $5.3\pm3.2$ & $7.0$\\
		
		\hline
		\multirow{3}{2.2cm}{Current work} 
		
		& Cardiac CTA & Left ventricle & 50 & 50 & $3.0\pm3.7$ & $4.5\pm3.4$ & $2.5$\\
		& Abdomen CT & Liver & 50 & 49 & $\phantom{1}8.9\pm15.0$ & $16.9\pm11.5$ & $6.4$\\
		& Chest CT & 4 structures & 100 & 100 & $3.4\pm4.1$ & $5.3\pm3.7$ &  $1.7$\\
	\end{tabular}
\end{table*}
\subsection{Comparison with Other Methods}
Previous work presented localization methods for different anatomical structures in different types of images and frequently used different evaluation metrics. In addition, the definition of the localization protocol was not always provided and might have been different than used in this study. Hence, it is difficult to reliably compare performance of different methods. Nevertheless, reported results can be used as estimates and used to indicate differences among the methods. Table~\ref{tab:comparison_distance} lists results as reported in the original works. 

To enable direct comparison, our preliminary work\cite{devos2016} was evaluated on the datasets used in the current work. The current method shows better results for localization tasks in cardiac CTA and abdomen CT. Similar, but much faster, localization results were achieved for the tasks in chest CT.

\section{Discussion}
\label{sec:discussion}
A method for localization of single or multiple anatomical structures in 3D images using ConvNets has been presented. \textit{Localization} in 3D volumes is achieved through \textit{detection} of presence of the target structure(s) in all 2D image slices of a 3D volume. 

In localization of single anatomical structures, the best results were achieved when contrast with the neighboring tissue was clear. This can be seen e.g. in accurate localization of the heart in sagittal slices where the heart borders the lungs. In cases when contrast was low or the task definition was less clearly defined, e.g. in localization of the inferior and sinister boundaries of the liver, the results of the method showed larger variability. This variability is also shown in the analysis of differences between observers; these results follow the same trend as those of the proposed automatic method. The method achieved similar localization results for multiple anatomical structures. Localization of multiple or single structures did not fail in any of the scans. However, localization of multiple structures showed less severe outliers. Because localization of single anatomical target structures is a less complex problem than localization of multiple target structures and the same network has been used in both applications, the learned features were probably more generalizable for the complex problems, which results in higher robustness.

With the aforementioned reasoning one might assume that training one ConvNet for all localization tasks combined might further improve robustness. Nevertheless, this might be compromised by conflicting localization tasks. The possibility exists that anatomical structures of interest (e.g. the heart) appear in image slices of a dataset where the structure is not defined as a target (e.g. axial image slices from abdomen CT). The method learns to detect presence of the structure in one dataset, but learns to ignore it in the other dataset.

On the other hand, automatic localization might be further improved by substantially enlarging the dataset, especially given that the ConvNets are known to perform well when lots of training data are available. However, obtaining large sets of annotated training data in medical images is typically cumbersome and sometimes even impossible. For localization purposes, the here presented accuracy is likely sufficient when used as a preprocessing step for segmentation or for image retrieval. The presented results have been obtained with relatively small training sets where setting the reference standard has been fast and simple. Utilization of small training sets may allow easy application of the method for localization of other anatomical structures. 

In the presented work, the ConvNet used for \textit{localization} in 3D images was selected based on its \textit{detection} performance in 2D image slices, because localization performance is directly dependent on correct detection. Different networks were evaluated using detection of the left ventricle in image slices from cardiac CTA. This task was chosen because of its complexity. Namely, cardiac CTA scans have a substantially variable field of view, and moreover, the left ventricle has boundaries that at places have to be inferred as well as boundaries that are clearly defined. The best detection performance was achieved by BoBNet and thus it was used for localization.

In the presented method, spatial dependencies among different anatomical structures were not specified. Although the 2D ConvNet finds these spatial relations in higher abstraction layers, the proposed method could benefit from contextual information provided by inputs of other anatomical structures of interest found in image slices and view planes. Alternatively, this could be achieved with recurrent neural networks  (RNNs)~\cite{hochreiter_long_1997} that find relations in sequences. In future work, we will investigate training an RNN with sequences of adjacent slices, which might provide information about the relation among adjacent slices. This may further improve localization performance.

Previously, ConvNets have been used to locate 2D objects, e.g. by using a sliding window detector or using region proposals as in regions with CNN features (R-CNN)\cite{girshick2014}. Extension of these methods to 3D medical images would not be trivial. First, using a sliding window in 3D would exponentially increase the number of windows to be analyzed. Second, the region proposals in R-CNN are generated by the selective search method described by Uijlings et al.~\cite{uijlings2013}, which is a hierarchical unsupervised segmentation method. Such segmentation methods perform well in 2D natural images with low noise levels, high contrast between structures, and a three channel RGB color space, but might be less suited for application in images with high noise levels, low contrast between tissues, and single channel images. Furthermore, the number of regions that would need to be analyzed would probably be high and thereby the computation time would increase dramatically. Moreover, analysis of 3D images would possibly require more training data. It is likely that with the proposed approach where 3D bounding boxes are created by detecting anatomical target structures in the lower 2D dimension, the low number of necessary training images is optimally exploited.

Compared to our preliminary work, the current method shows better localization results in cardiac CTA and abdomen CT. Even though the preliminary method was specifically developed for localization in chest CT, the current method achieved similar localization results in chest CT but substantially faster. The results demonstrate robustness of the method in a diverse set of images with respect to acquisition parameters (see Table~\ref{tab:data}). Furthermore, a direct comparison of performance with previously published algorithms cannot be made, because different datasets and different task definitions were applied (e.g. rotated bounding boxes vs. bounding boxes aligned with the image planes). Nevertheless, the obtained results demonstrate that the proposed method achieved competitive results with respect to accuracy and processing time. Furthermore, we would like to stress that no postprocessing was performed to the ConvNet outputs. Postprocessing such as smoothing of the ConvNet responses might further increase performance. Also, the processing time of the method could be further reduced by parallel instead of sequential analysis of image slices.

Advanced multi-atlas based methods might be able to achieve similar performance to the proposed approach. However, besides fast processing times of the proposed approach, its main advantage  is the  ability to be built into a convolutional neural network for e.g. (subsequent) segmentation task. The network could then  be trained in an  end-to-end fashion allowing optimal training for the task at hand.

Finally, the method only requires a training set that contains bounding boxes parallel to the viewing planes. This means that generating training data by an expert is simple and fast: neither principal axes of anatomical structures have to be determined, nor rotational alignment of bounding boxes have to be performed. Merely an indication of the location of anatomical structures has to be provided by outlining the full anatomical structure of interest with a bounding box. The method takes extracted image slices as its input and it does not require any task specific information. Hence, it could be straightforwardly applied for localization of other anatomical structures in other imaging modalities. 

\section{Conclusion}
\label{sec:conclusion}
A method for localization of single or multiple anatomical structures of interest in 3D medical images using a ConvNet has been presented. The method can localize one or more anatomical target structures robustly, with high speed, and high accuracy, using only a limited number of training examples.

\ifCLASSOPTIONcaptionsoff
  \newpage
\fi




\IEEEtriggeratref{19}
\bibliographystyle{IEEEtr_bob}
\bibliography{bibliography}

\end{document}